\documentclass[conference]{IEEEtran} 
\let\labelindent\relax
\usepackage{graphicx}
\usepackage{type1cm}
\usepackage{lettrine}
\usepackage{amsmath,amsthm,amssymb}
\usepackage{amsfonts}
\usepackage{enumitem}
\usepackage{subcaption} 
\usepackage{rotating}
\usepackage{url}
\usepackage{algorithm}
\usepackage{algorithmic}
\usepackage[numbers]{natbib}
\usepackage{multicol}
\usepackage[bookmarks=true]{hyperref}
\usepackage{xcolor}

\begin{document}

\title{Accelerated RRT* By Local Directional Visibility}

\author{Chenxi Feng$^{*}$, Haochen Wu$^{*}$
\\ University of Michigan, Ann Arbor, MI
\\ \{chenxif, haochenw\}@umich.edu
\\$^{*}$Equal Contribution}

\maketitle
\IEEEpeerreviewmaketitle
\textbf{\textit{Abstract -}}
\textbf{RRT* is an efficient sampling-based motion planning algorithm. However, without taking advantages of accessible environment information, sampling-based algorithms usually result in sampling failures, generate useless nodes, and/or fail in exploring narrow passages. For this paper, in order to better utilize environment information and further improve searching efficiency, we proposed a novel approach to improve RRT* by 1) quantifying local knowledge of the obstacle configurations during neighbour rewiring in terms of directional visibility, 2) collecting environment information during searching, and 3) changing the sampling strategy biasing toward near-obstacle nodes after the first solution found. The proposed algorithm RRT* by Local Directional Visibility (RRT*-LDV) better utilizes local known information and innovates a weighted sampling strategy. The accelerated RRT*-LDV outperforms RRT* in convergence rate and success rate of finding narrow passages. A high Degree-Of-Freedom scenario is also experimented.}

\section{Introduction}
\lettrine{M}{otion} planning plays a key part in robot execution and environment recognition. Many approaches have been proposed, from discrete map searching to random sampling searching. Due to less computational complexity in high-dimensional space and no explicit obstacle information required, random sampling methods gained popularity in recent years. Stochastic searching methods, such as Rapidly-exploring Random Trees (RRTs) \cite{RRT}, Probabilistic Roadmaps (PRMs) \cite{PRM}, and Expansive Space Trees (ESTs) \cite{EST}, use sampling-based methods to avoid discretization of the state space as a requirement. This allows them to scale more effectively with problem size and to directly consider kinodynamic constraints; however,the drawback is a less-strict completeness guarantee. RRTs are probabilistically complete, guaranteeing that the probability of finding the optimal solution, if one exists, approaches unity as the number of iterations approaches infinity. RRT is an efficient obstacle free path finding algorithm, where smoothing strategies \cite{smooth} could be further executed to find a smoother and shorter path. Although it ensures probabilistic completeness, it cannot guarantee finding the most optimal path. RRT*\cite{RRT*} is an incremental sampling based algorithm to approach an optimal solution ensuring asymptotic optimality, but it has been proven mathematically that it reaches the optimal solution in infinite time. RRT* is the most suitable for single-query or dynamic planning and able to explore the complex environment efficiently.

People learn from their experiences, but RRT* \cite{RRT*} planner does not. RRT* randomly takes samples around the configuration space, but the sampling strategy never improves. After hundreds or even thousands sampling iterations, the planner still samples blindly in the space. Every time the robot visits a location, robots should be able to collect some environment information and learn from its failures. RRT* ensures probabilistic completeness and asymptotic optimality, but it does not take prior knowledge of the environment into account, generating many useless nodes and wasting the computation power. 

\begin{figure}[b]%
    \centering
    \begin{subfigure}{0.153\textwidth}
        \centering
        \includegraphics[width=\textwidth]{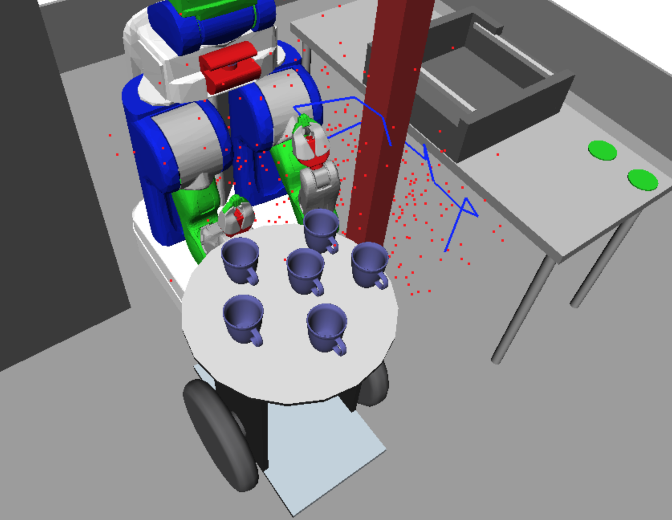}
        \caption{ }
    \end{subfigure}%
    \hspace{1mm}
    \begin{subfigure}{0.153\textwidth}
        \centering
        \includegraphics[width=\textwidth]{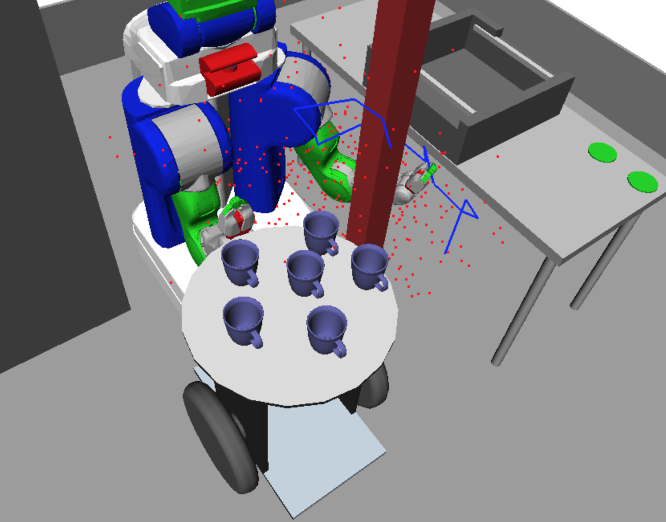}
        \caption{ }
    \end{subfigure}
    \hspace{0mm}
    \begin{subfigure}{0.152\textwidth}
        \centering
        \includegraphics[width=\textwidth]{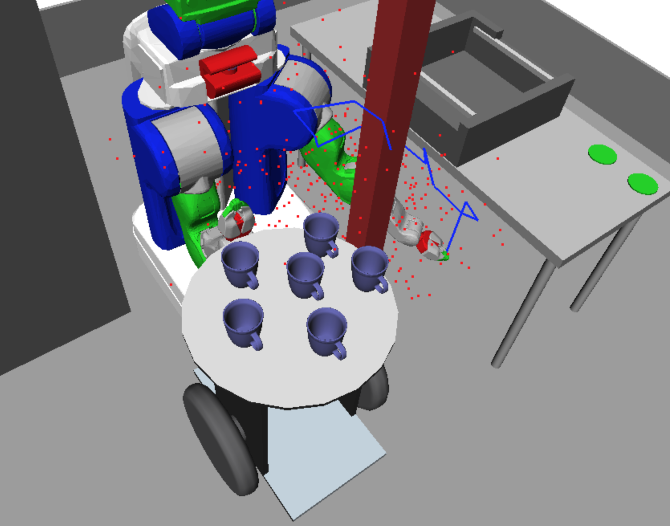}
        \caption{ }
    \end{subfigure}
        \caption{Sequence of the snapshots of the execution with PR2 robot in high DOF. Red dots are the near-obstacle ndoes.}
    \label{fig:hd123}
\end{figure}
\textit{Contributions:} The proposed algorithm, RRT*-Local Directional Visibility (RRT*-LDV), is an extension of RRT*. The proposed RRT*-LDV is implemented and provided as an open-source at \href{https://github.com/chenaaron0917/RRTstar-LDV}{\textit{github.com/hcaawu/RRTstar-LDV}}. An example of solution path found in high DOF is demonstrated in Fig. \ref{fig:hd123}, where red dots indicate the end-effect of the robot arm in the near-obstacle node configurations and the blue line is the end-effect of the best trajectory to the final configuration. To ensure faster convergence and find narrow passages more easily, RRT*-LDV extends RRT* with:
\begin{enumerate}
    \item A strategy to quantify local information by defining directional visibility for all nodes (Section \ref{DefLDV})
    \item An approach to collect environment information while searching by creating a set of near-obstacle nodes (Section \ref{Xfail})
    \item A intelligent sampling strategy to decide where to sample biasing toward near-obstacle nodes (Section \ref{Imp})
\end{enumerate}

The rest of the paper is outlined as follows: Section \ref{litreview} reviews other extensions of RRT* and different approaches in sampling-based algorithms. Section \ref{method} formulates the problem and explains the proposed approach along with the algorithms. Results and statistical analysis are presented in Section \ref{results}. Finally, Section \ref{conclusion} summaries the proposed algorithm and discusses the results and future works.

\section{Related Work}
\label{litreview}
Prior work focus has relied on two methods to increase the convergence rate: 1)save the computational cost 2)propose a novel sampling strategy. In the first category of methods, \cite{AOP} presents an implementation of RRT* with delayed collision checks by reducing the computational effort cost on checking the validity of the new vertex and its connection with other nodes to improve the performance, which sorts the nearby candidates and then checks from the most likely one instead of all the vertices. In the second category, plenty of nonuniform sampling strategies are introduced to the RRT-based algorithms based on different types of knowledge being utilized. The biased sampling methods can usually be divided into different types, e.g., goal-biased, path-biased, and obstacle-based sampling.

\subsection{Sample Biasing}
\subsubsection{Heuristic-biased Sampling:} Heuristic-biased sampling attempts to increase the probability of sampling certain configurations by weighting the sampling of the configuration space with a heuristic estimate of each state. It is used to improve the quality of a regular RRT such as selecting states with a probability inversely proportional to their heuristic cost by Urmson and Simmons \cite{hRRT} in the Heuristically Guided RRT (hRRT). The hRRT was shown to find better solutions than RRT; however, the use of RRTs means that the solution is almost surely suboptimal \cite{hRRT}. Kiesel et al \cite{fbias} uses a two-stage process to create an RRT* heuristic in their f-biasing technique. 

\subsubsection{Path biased sampling} Path-biased sampling attempts to increase the frequency of sampling around the current solution path. This approach assumes that the current solution is either homotopic to the optimum or separated only by small obstacles. As this assumption is not generally true, path-biasing algorithms must also continue to sample globally to avoid local optima. The ratio of these two sampling methods is frequently a user-tuned parameter. Alterovitz et al. \cite{RRM} use path biasing to develop the Rapidly-exploring Roadmap (RRM). Once an initial solution is found, each iteration of the RRM either samples a new state or selects an existing state from the current solution and refines it. Path refinement occurs by connecting the selected state to its neighbours resulting in a graph instead of a tree.

\subsubsection{Obstacle-based sampling} Obstacle-based sampling is the most popular one among these methods. It takes advantage of prior knowledge of the known obstacles or invalidity of previous samples to guide the next (or future) sampling \cite{OB,EEB}. A large amount of works have been committed to propose better utilizing local and map information such as Informed RRT* \cite{Informed}.Different methods are also applied to rejection sampling to simplify the tree in less important areas \cite{ATD}.Among them, Batch Informed Trees (BIT*)\cite{BIT},by searching the subset of state space,speed up the convergence and improve the quality of solution. One of its extension:Regional Accelerated Batch Informed Trees\cite{RABIT} extend BIT* by using optimization to exploit local domain information and find alternative connections for edges in collision and accelerate the search. This improves search performance in problems with difficult-to-sample homotopy classes (e.g., narrow passages) while maintaining almost-sure asymptotic convergence to the global optimum. Besides, RRT*-Smart improves the initial solution by sampling around "beacons" to make the solution converge to optimal solution fast \cite{Smart}.

\subsection{Intelligent Sampling} 
Ichter et al. \cite{LSD} added offline learning phase to update sampling distribution based on successful planning and demonstrations. However, learning-based planning method requires a large amount of successful path data or human demonstration in the training phase. SL-RRT* \cite{SL} by Perez et al. extend the sampling scheme of RRT* by introducing a self-learning mechanism with prior knowledge maintenance and hybrid-biased sampler. This mechanism requires large computation efforts because it computes the visibility in every direction of the node.

Our proposed work falls in the scope of obstacle-based sampling with self-awareness ability. The RRT*-LDV algorithm randomly take samples in the configuration space as RRT* does before the first solution is found. While searching, a empty set called near-obstacle nodes will be initialized and collects nodes close to obstacles. Once this first path is found, it then optimizes it by sampling around the near-obstacle nodes. This optimized path yields biasing points around the near-obstacle nodes. Furthermore, among all these near-obstacle nodes, some are worth to be sampled than others and the near-obstacle node selection will be biased by their importance. The details of the proposed approach would be discussed in Section \ref{method}.

\section{Methodology}
\label{method}
\subsection{Problem Statement}
Let $X$ define the whole configuration space, $X_{goal}$ define a goal region, and $X_{obs}$ define the obstacle region or the region where the robot violates the system constraints $\mathcal{C}$. The obstacle-free space is then $X_{free} = X \setminus X_{obs}$. Given the initial configuration $x_{start} \in \mathbb{R}^d$, $X_{goal}$, and $X_{obs}$ for the collision checker, the goal of the problem is to solve a sequence of control inputs $u[0:T] \in \mathcal{U}$ which moves the robot to follow a path $x(t) \in X_{free}$ that starts from $x(0)=x_{start}$ to $x(T) \in X_{goal}$, and the path would be further optimized after the first solution is found.

During searching, the proposed algorithm RRT*-LDV grows a tree $G=(V,E)$ of nodes $V \in X_{free}$ and edges $E$ connecting these nodes. The node $\{(x,vis)\} \in V$ contain not only the configuration $x$ but also local directional visibility $vis$ which would be defined and discussed in Section \ref{DefLDV}. In addition, a set of near-obstacle nodes $X_{fail}$ would be collected for weighted sampling (Section \ref{Xfail}).
\subsection{Approach Overview}
The proposed RRT*-LDV presented in Algorithm \ref{algo1} tries to develop a intelligent sampling strategy that utilizes accessible environment information to solve the above motion planning problem. It follows the same structure of extending the search tree as RRT*. Lines 6,7,9,11 to 14 are the default steps of RRT*, which are briefly explained as follows:
\begin{itemize}
    \item \textbf{Nearest Neighbor:} The function $nearest$ returns the nearest node in the tree $G$ to $x_{rand}$ in terms of the euclidean distance.
    \item \textbf{Steering:} This function finds a candidate node $x_{new}$ based on steer size $\eta$, which is to solve $steer(x,y)=argmin_{z \in R^d,||z-x||\leq \eta}||z-y||$.
    \item \textbf{Check Collision:} The function $collisionFree$ determines if the candidate node $x_{new}$ satisfies the system constraints $\mathcal{C}$ and lies in the obstacle-free space $X_{free}$.
    \item \textbf{Near Neighbors:} This function returns a set of vertices $X_{near} \in V$ within a closed ball of radius $r_n$ centered at $x_{new}$, where $r_n = min\{(\frac{\gamma}{\xi_d}\frac{log(n)}{n})^{1/d},\eta\}$. $\gamma$ is a constant, $\xi_d$ is the volume of a unit ball in $\mathbb{R}^d$, $n$ is the number of vertices in the tree.
    \item \textbf{Best Parent:} The function $bestParent$ finds the best parent $x_{min} \in X_{near}$ for $x_{new}$ based on the cost-to-come heuristic.
    \item \textbf{Rewiring:} $rewire$ occurs if the heuristic cost of the nodes in $x_{near} \in X_{near}$ is less through $x_{new}$ than the current cost and set the parent of $x_{near}$ to $x_{new}$
\end{itemize}

\begin{algorithm}[t]
\caption{RRT*-LDV}
\hspace*{\algorithmicindent} \textbf{Inputs}: $x_{start}$, $X_{goal}$, $X_{obs}$
\begin{algorithmic}[1]
\label{algo1}
\STATE {\color{red}$V \leftarrow (x_{start},vis=\infty);$} $\; E \leftarrow \emptyset$
\STATE $G \leftarrow initTree(V,E)$
\STATE {\color{red} $X_{fail} \leftarrow \emptyset; \; firstFound \leftarrow False$}
\WHILE{$i \leq maxIter$}
\STATE {\color{red} $x_{rand} \leftarrow sampleAroundXfail(i,X_{fail})$}
\STATE $x_{nearest} \leftarrow nearest(G,x_{rand})$
\STATE $x_{new} \leftarrow steer(x_{nearest},x_{rand})$
\STATE {\color{red} $X_{fail} \leftarrow addXfail(x_{nearest},x_{new},X_{fail})$}
\IF{$collisionFree(x_{new},x_{nearest},\mathcal{C})$}
\STATE {\color{red}$V' \leftarrow V \cup \{(x_{new},vis=\infty)\}$}
\STATE $X_{near} \leftarrow nearNeighbors(G,x_{new})$
\STATE $x_{min} \leftarrow bestParent(X_{near},x_{new})$
\STATE $E' \leftarrow E \cup \{(x_{min},x_{new})\}$
\STATE $E \leftarrow rewire(X_{near} \setminus \{x_{min}\},x_{new},E')$
\STATE {\color{red} $V \leftarrow updateVisibility(X_{near},x_{new},V')$}
\STATE {\color{red} $X_{fail} \leftarrow updateImportance(X_{fail},V)$}
\STATE $G \leftarrow (V,E)$
\ENDIF
\ENDWHILE
\end{algorithmic}
\end{algorithm}

The contributions of this work are implemented in the highlighted portion. $updateVisibility$ (line 15) quantifies and stores local directional visibility for each node (Section \ref{DefLDV}). $addXfail$ (line 8) collects near-obstacle nodes for sampling (Section \ref{Xfail}). The proposed new sampling strategy is implemented in $sampleAroundXfail$ (line 5) which selects a near-obstacle node and samples around it for faster convergence and narrow passage detection. $updateImportance$ updates the importance of each near-obstacle node for sampling selection (Section \ref{Imp}).

\subsection{Local Directional Visibility}
\label{DefLDV}
\begin{figure}[t]%
    \centering
    \includegraphics[width=0.4\textwidth]{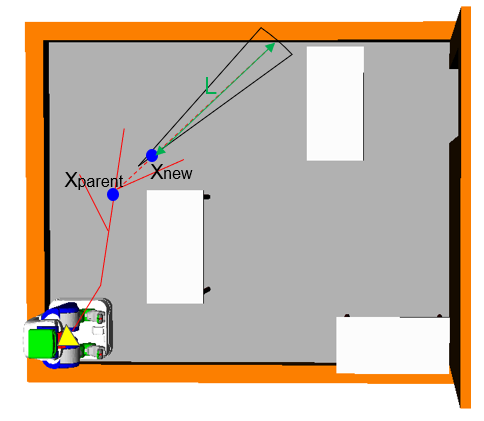}
    \caption{When a new node $X_{new}$ is added to the tree and $x_{parent}$ is chosen to be its parent, then the directional visibility is represented by the green arrow line whose length is $L$.}
    \label{fig:LDV}
\end{figure}

Local directional visibility is a measurement of distance from a specific node to obstacles in a certain direction. Every time a node $x_{new}$ is initialized and added to the node tree, the direction of this new branch is from the parent node to the new node, and the unit direction is  $dir = \frac{x_{new} -x_{parent}}{||x_{parent},x_{new}||}$ as shown in Fig. \ref{fig:LDV}. The unit direction is assigned as a property to $x_{new}$ called $dir$. By extending this new edge from $x_{new}$ along the $dir$, it will either hit some obstacles or exceed the working space limit after a visible distance $L$. $L$ implies how far $x_{new}$ is from the obstacles in $dir$ and will be assigned as a property called local directional visibility $vis$ to $x_{new}$. Once the parent of the node is changed during the $bestParent$ or $rewiring$ step, the $dir$ and $vis$ of the corresponding nodes will be updated accordingly. 

Hence, the Local Directional Visibility is computed as:
\begin{equation}
vis=argmin_{L \in \mathbb{R}, x_{obs}\in X_{obs}} ||x+dir*l-x_{obs}||
\end{equation}
where again $L$ is the visible distance to reach $X_{obs}$ in the direction $dir$. 
This visibility property is calculated whenever a node is added to the tree. It will be utilized when calculating the importance of the near-obstacle node in Section \ref{Imp}. This visibility quantification is local and directional because it only considers a single direction from the parent of the node to the node.

RRT* chooses the best parent node with the lowest cost to come. It is worth to mention that we tried another algorithm, which selects the parent node with lower cost to come and higher visibility. We expected to see that algorithm will spend less time finding the first solution. However, that algorithm did not perform very well. In that algorithm, the tree branch “bends backwards” at the boundary of obstacles. It implies that the new node tends to connect to a parent node which gives it a higher local directional visibility. That algorithm helped to make the tree grow towards the areas with high visibility in the local sense but not in global sense. So, the time spent of finding the first solution is almost the same as the original RRT*.
\subsection{Near-Obstacle Nodes}
\label{Xfail}
During searching, environment information could be easily collected and utilized for sampling strategies. The most important environment information is the configuration of obstacles. If this information is known to the planner, the robot would not waste time on sampling useless configurations that neither improve the current solution nor lie in $X_{free}$. Hence, we proposed an approach to collect a set of near-obstacle nodes $X_{fail}=\{x \in \mathbb{R}^d, ||x-x_{obs}||\leq \eta,  \forall x_{obs} \in X_{obs}\}$. The distance from every near-obstacle nodes to obstacle configurations is at most the steer size $\eta$. The set of these near-obstacle nodes is served as known knowledge to the robot for intelligent weight sampling.
\begin{algorithm}[h]
\caption{addXfail($x_{nearest},x_{new},X_{fail}$)}
\begin{algorithmic}[1]
\label{algo2}
\STATE $(x_{fail},imp=0) \leftarrow getxfail(x_{nearest},x_{new})$
\IF{$X_{fail}$ is empty}
\STATE $X_{fail} \leftarrow X_{fail} \cup \{(x_{fail},imp)\}$
\ELSIF{$nearest(X_{fail},x_{fail})\geq \rho_{fail}$}
\STATE $X_{fail} \leftarrow X_{fail} \cup \{(x_{fail},imp)\}$
\ENDIF
\RETURN $X_{fail}$
\end{algorithmic}
\end{algorithm}

To ensure $X_{fail}$ grows fast enough and to prevent the size of $X_{fail}$ from growing infinitely and taking too much computational time, two requirements are added and local directions are utilized. First, we tend to add a near-obstacle node every time after steering and follow the steering direction $dir$ from $x_{nearest}$ to $x_{new}$, which makes the set grow fast. The configuration of the near-obstacle node can be computed in $getxfail$ (Algorithm \ref{algo2} line 1) as:
\begin{equation}
    x_{fail}=x_{new}+s*dir(x_{nearest},x_{new})
\end{equation}
where $s$ is one step less than the number of steering steps to reach $X_{obs}$ or fail the system constraints $\mathcal{C}$. Second, we define a density $\rho_{fail}$ for the near-obstacle nodes in the environment. The near-obstacle node would only be added to the set if the set is empty (Algorithm \ref{algo2} line 2) or the set is not dense enough (Algorithm \ref{algo2} line 4).

\begin{figure}[t]%
    \centering
    \begin{subfigure}{0.23\textwidth}
        \centering
        \includegraphics[width=\textwidth]{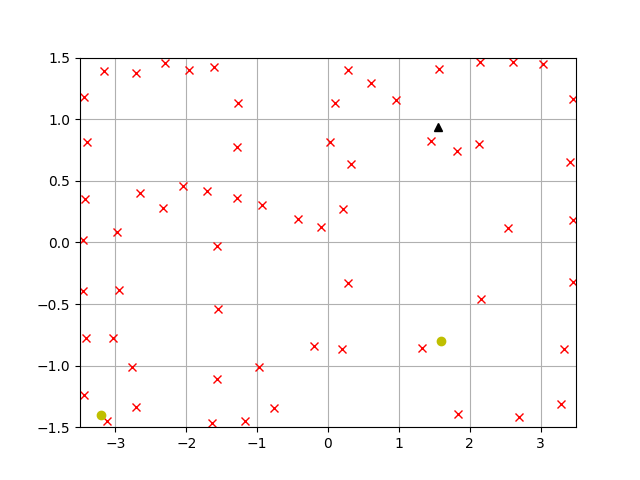}
        \caption{ }
    \end{subfigure}%
    \hfill
    \begin{subfigure}{0.23\textwidth}
        \centering
        \includegraphics[width=\textwidth]{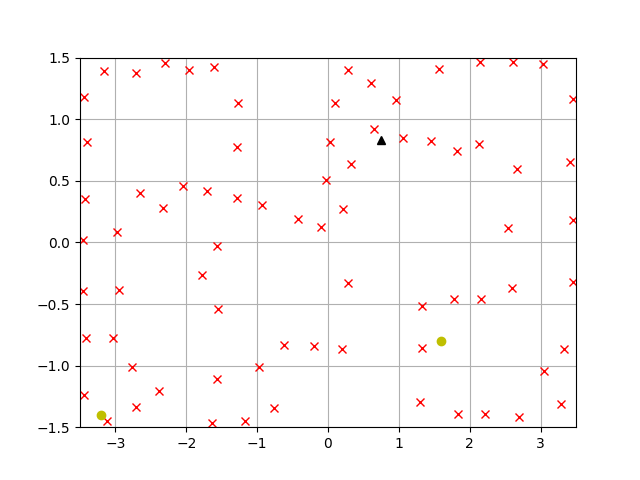}
        \caption{ }
    \end{subfigure}
        \caption{Visualization of near-obstacle nodes set: (a) after the first solution, (b) after 1500 iterations. Red dots are near-obstacle nodes and the circles indicate the potential areas to perform sampling}
    \label{fig:failnodes}
\end{figure}

Once the first solution is found, the set of near-obstacle nodes is large as shown in Fig. \ref{fig:failnodes}(a), which is desired for the proposed sampling strategy in Section \ref{Imp}. Red crosses are the near-obstacles nodes.

\subsection{Weighted Sampling Strategy}
\label{Imp}
Before the first solution is found, RRT*-LDV algorithm inherits the same sampling strategy as RRT* (Algorithm \ref{algo3} line 1,2). Throughout the planning, RRT*-LDV increases local knowledge by updating local directional visibility, and stores environment information by collecting a set of near-obstacle nodes. Now, to utilize the these two types of information, a new weighted sampling strategy is proposed biasing towards near-obstacle nodes $X_{fail}$ and areas with high local directional visibility and low number of nodes. This strategy has two folds  (Algorithm \ref{algo3} line 4, 5): (1) sampling around near-obstacle nodes and (2) selecting the most important near-obstacle node for sampling.
\begin{algorithm}
\caption{sampleAroundXfail($i,X_{fail}$)}
\begin{algorithmic}[1]
\label{algo3}
\IF{$fisrtFound$}
\RETURN $x_{rand} = sample(i)$
\ELSE
\STATE $x_{fail} \leftarrow selectMostImpFailx(X_{fail})$
\RETURN $x_{rand} = sampleAround(x_{fail})$
\ENDIF
\end{algorithmic}
\end{algorithm}

\begin{figure}[b]%
    \centering
    \begin{subfigure}{0.22\textwidth}
        \centering
        \includegraphics[width=\textwidth]{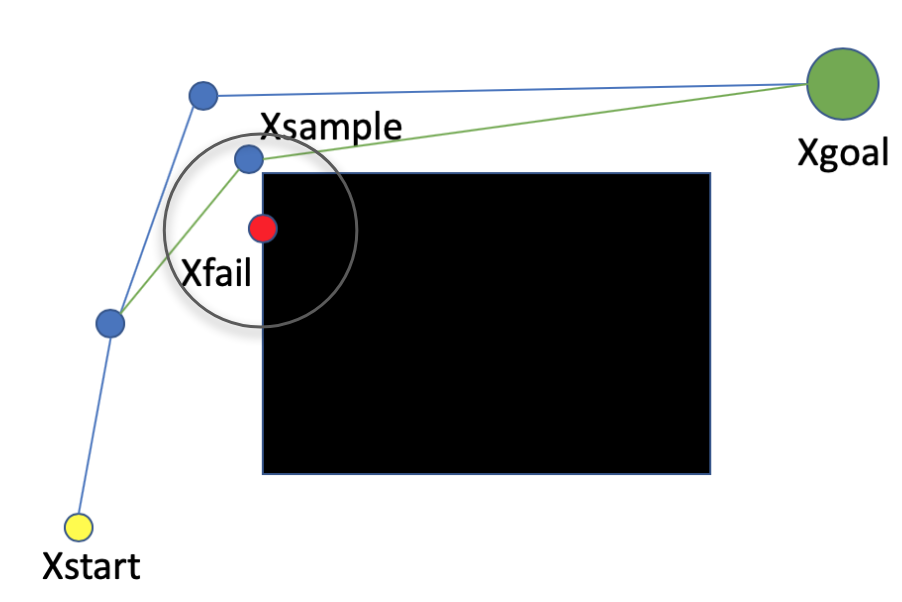}
        \caption{ }
    \end{subfigure}%
    \hfill
    \begin{subfigure}{0.18\textwidth}
        \centering
        \includegraphics[width=\textwidth]{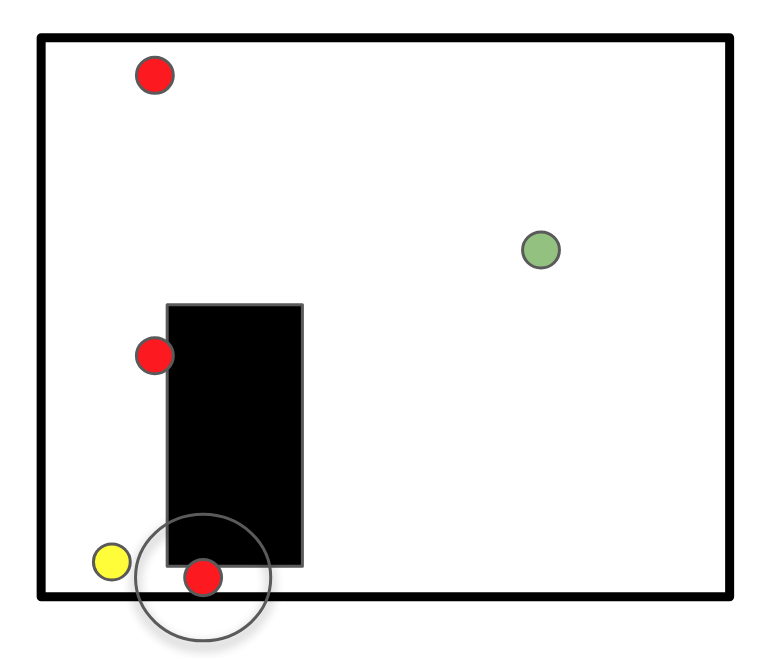}
        \caption{ }
    \end{subfigure}
        \caption{Advantages of sampling around near-obstacle nodes: (a) close to obstacles' vertices, (b) find narrow passages}
    \label{fig:xfailadv}
\end{figure}
\begin{figure*}[h]%
    \centering
    \begin{subfigure}{0.3\textwidth}
        \centering
        \includegraphics[width=\textwidth]{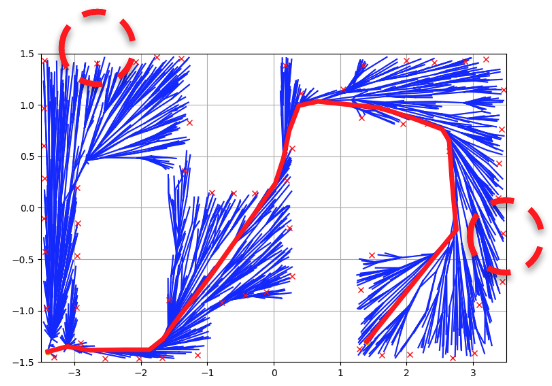}
        \caption{$x_{fail}$ along obstacle flat edges}
    \end{subfigure}%
    \hfill
    \begin{subfigure}{0.3\textwidth}
        \centering
        \includegraphics[width=\textwidth]{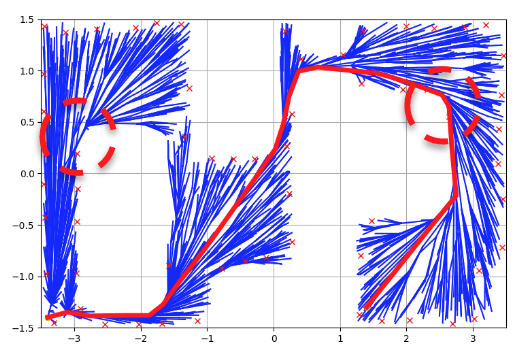}
        \caption{$x_{fail}$ around obstacle vertices}
    \end{subfigure}
    \hfill
    \begin{subfigure}{0.3\textwidth}
        \centering
        \includegraphics[width=\textwidth]{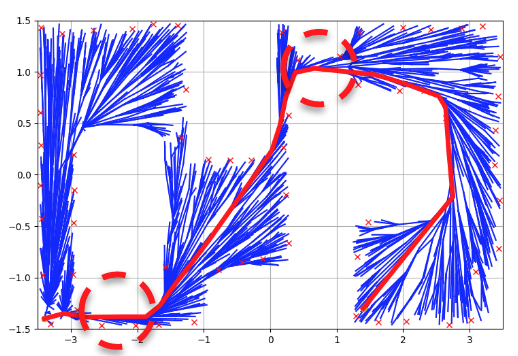}
        \caption{$x_{fail}$ in narrow passages}
    \end{subfigure}
        \caption{Demonstration of the importance of near-obstacle nodes in terms of local direction visibility and number of nearby nodes. (a) low visibility, large number of nodes, (b) high visibility, large number of nodes, (c) relative high visibility, small number of nodes.}
    \label{fig:xfailimp}
\end{figure*}
\subsubsection{Sampling Around Near-Obstacle Nodes} There are two major advantages of sampling around $X_{fail}$ as shown in Fig. \ref{fig:xfailadv}. Since it is known that the optimal path converges to obstacle vertices. Many $X_{fail}$ collected reside near the obstacle vertices. After the first solution is found, a sampling bias $\lambda_s$ is put on the $X_{fail}$ to let the planner have more chances to sample around them for faster convergence and ensure probabilistic completeness. Besides, some $X_{fail}$ may reside near/in narrow passages as shown in Fig. \ref{fig:xfailadv}(b). Sampling around them would help the planner to find narrow passages more easily.

Since all the near-obstacle nodes are located near obstacles with at most distance of steer size $\eta$. It is recommended to sample with in the $d$-dimensional cube region around $X_{fail}$.

\subsubsection{Near-Obstacle Node Importance}
Sampling around a randomly picked near-obstacle node still wastes a lot of time on those $x_{fail}$ that can not improve the path. To further improve the sampling strategy, it is desired to determine which near-obstacle nodes are more important in the ways that they are close to obstacles' vertices and near or in narrow passages so that the most important $x_{fail}$ would be chosen for sampling.

To achieve this, RRT*-LDV takes advantages of the local directional visibility $vis$ of each node in the tree. $X_{fail}$ could be categorized as three types as shown in Fig. \ref{fig:xfailimp}:
\begin{enumerate}[label=(\alph*)]
    \item For $x_{fail}$ along obstacle flat edges, there are many nodes in the closed ball area around $x_{fail}$ and they have low $vis$ because they point into the obstacles. Thees nodes are not desired for selection because they do not improve the solution or find narrow passages.
    \item For $x_{fail}$ around obstacle vertices, they demand more exploration because the shortest path converges to the vertices. The number of nodes in the closed ball area is large but they have high $vis$ because they point into the free space.
    \item For $x_{fail}$ in narrow passages, they demand more exploration because they might lead to shorter path. The number of nodes in the closed ball is very small but they have relative high $vis$.
\end{enumerate}

Therefore, to define the importance of each $x_{fail}$ describing the demand for exploration, we first define a closed ball centered at each $x_{fail}$ with radius $r_f$: $\mathcal{B}_{f}=\{y\in \mathbb{R}^d, ||y-x_{fail}||\leq r_f\}$. Then the importance of each $x_{fail}$ defined as:
\begin{equation}
   Imp_{f}=\frac{\overline{vis}_{\mathcal{B}_f}}{(|\mathcal{B}_f|+1)^m}
\end{equation}
$\overline{vis}_{\mathcal{B}_f}$ is the average local directional visibility of nodes in $\mathcal{B}_f$. $|\mathcal{B}_f|$ is the number of nodes in $\mathcal{B}_f$. $m$ is a tuning parameter. Higher $m$ would bias the near-obstacle node selection toward the ones in narrow passages, while lower $m$ bias the selection toward the ones near obstacles' vertices. Once a near-obstacle node is selected and a sample is taken around it, the importance would decrease since the number of nodes in $\mathcal{B}_f$ increases, which ensures the chances for other near-obstacle nodes to be selected. A selection bias $\lambda_i$ is added to this strategy to make sure all near-obstacle nodes have chances to be selected.

\section{Results}
\label{results}
We implemented the proposed RRT*-LDV algorithm in Python interfacing with OpenRAVE, a C++ based environment for developing and deploying motion planning algorithms in robotics applications. All experiments were run on an Intel Core i7-8750H, 2.20GHz CPU with 15GB memory. Section \ref{2D} and \ref{highD} demonstrate the results for implementing in a simple two Degree-of-freedom scenario and a higher DOF scenario respectively. The statistical comparison between RRT* and RRT*-LDV is also presented in these sections.

\begin{figure}[b]%
    \centering
    \includegraphics[width=0.4\textwidth]{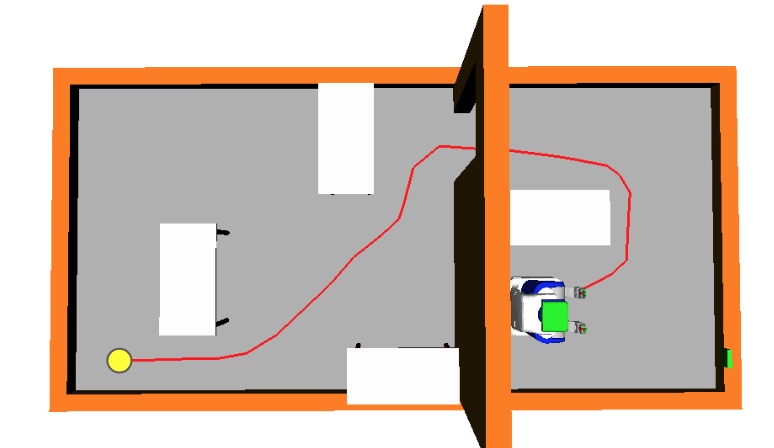}
    \caption{2DOF scenario. Yellow dot is the start configuration. Red line indicated the best path found after 1500 iterations.}
    \label{fig:env2}
\end{figure}
\subsection{2DOF Planning}
\label{2D}
A simple 2DOF (x,y coordinates) setting is considered where the robot is trying to find a collision-free path from the yellow start position to the final position as demonstrated in Fig. \ref{fig:env2}. The robot is expected to find the narrow passage going through the bottom which is the shorter path and path found should converge to the optional solution faster than RRT*.

\subsubsection{Node Tree Visualization}
\begin{figure}[t]%
    \centering
    \begin{subfigure}{0.16\textwidth}
        \centering
        \includegraphics[width=\textwidth]{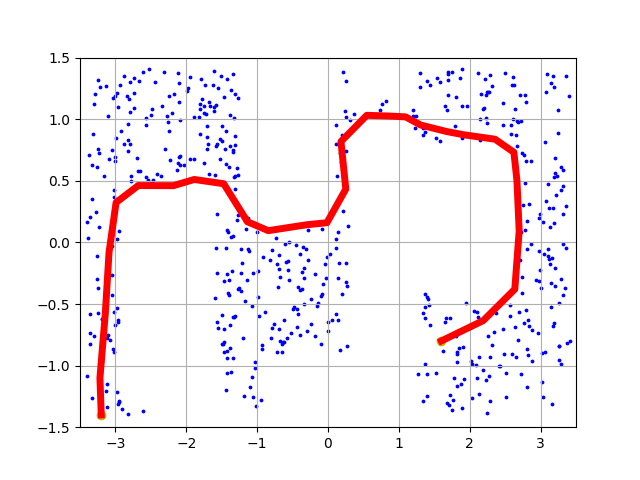}
        \caption{ }
    \end{subfigure}%
    \hfill
    \begin{subfigure}{0.16\textwidth}
        \centering
        \includegraphics[width=\textwidth]{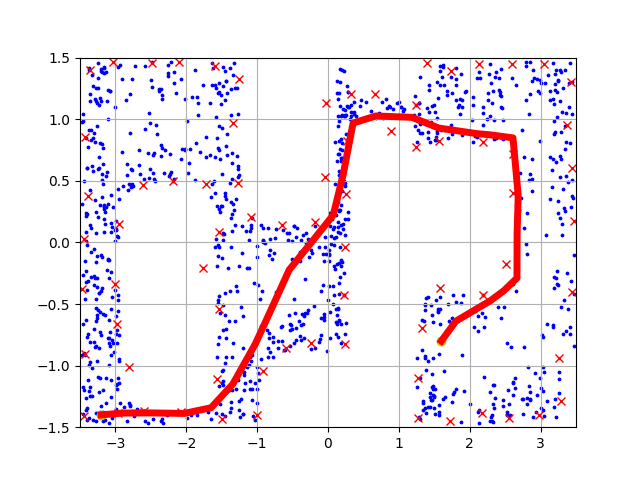}
        \caption{ }
    \end{subfigure}
    \hfill
    \begin{subfigure}{0.16\textwidth}
        \centering
        \includegraphics[width=\textwidth]{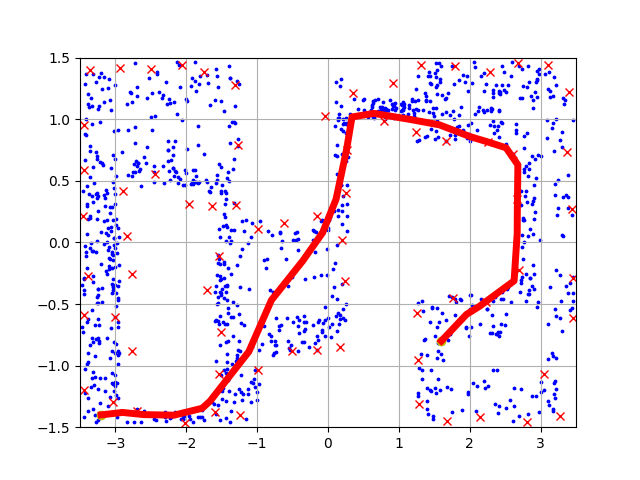}
        \caption{ }
    \end{subfigure}
        \caption{Tree nodes visualization. (a) RRT*, (b) Sampling around a randomly picked $x_{fail}$, (c) Sampling around the most important $x_{fail}$.}
    \label{fig:comp_nodetree}
\end{figure}

The visualization of the node tree under our sampling strategy compared to regular RRT* is presented in Fig. \ref{fig:comp_nodetree}. In (a), the tree nodes in regular RRT* is uniformly distributed; in (b) RRT*-LDV spends a lot of effort exploring the areas around near-obstacle nodes. However, sometimes exploration around certain near-obstacle nodes is not necessary; in (c), with weighted sampling strategy, RRT*-LDV is able to focus on exploring the areas around the more important near-obstacle nodes. The tree nodes is mainly distributed around obstacles' vertices and in narrow passages as expected.
\subsubsection{Evaluation and Analysis}
The proposed RRT*-LDV is evaluated the 2DOF scenario compared with the regular RRT*. To verify whether the new sampling strategy improves the first solution found faster than RRT*, we run the experiment for 1500 iterations and 20 times and compare their average cost and success rate of finding the narrow passage for each setting by varying sampling bias $\lambda_s$ and near-obstacle node selection bias $\lambda_i$.
\begin{figure}[b]%
    \centering
    \includegraphics[width=0.48\textwidth]{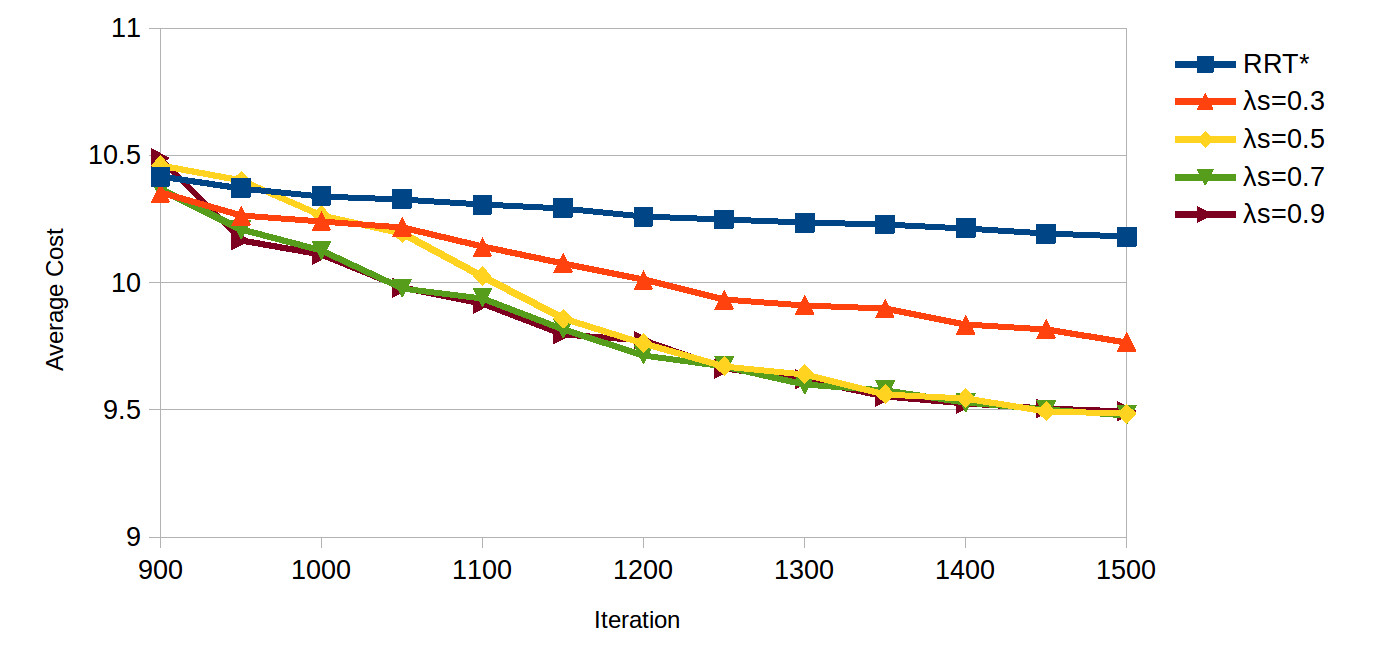}
    \caption{Sampling around a randomly picked near-obstacle node with increasing sampling bias $\lambda_s$}
    \label{fig:samplebias}
\end{figure}

Fig. \ref{fig:samplebias} presents the average path cost for RRT*-LDV with various $\lambda_s$ and no selection bias ($\lambda_s=0$) compared with RRT*. Higher $\lambda_s$ implies that the planner bias the sampling strategy more toward the areas around near-obstacle nodes $X_{fail}$. $\lambda_s=0$ indicates the near-obstacle node $x_{fail}$ is randomly selected. The path cost does decreases faster than RRT* and high $\lambda_s$ would converge to the optimal solution faster. Even half the sampling bias would provide significantly better result after 1500 iterations.

\begin{figure}[t]%
    \centering
    \begin{subfigure}{0.24\textwidth}
        \centering
        \includegraphics[width=\textwidth]{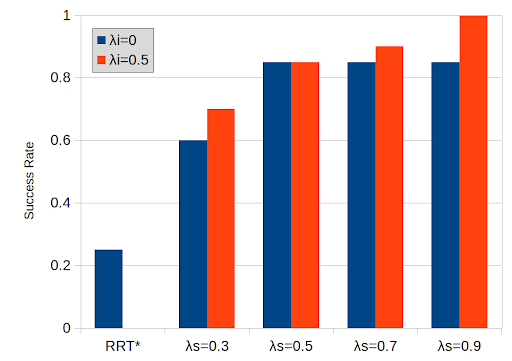}
        \caption{ }
    \end{subfigure}%
    \hfill
    \begin{subfigure}{0.24\textwidth}
        \centering
        \includegraphics[width=\textwidth]{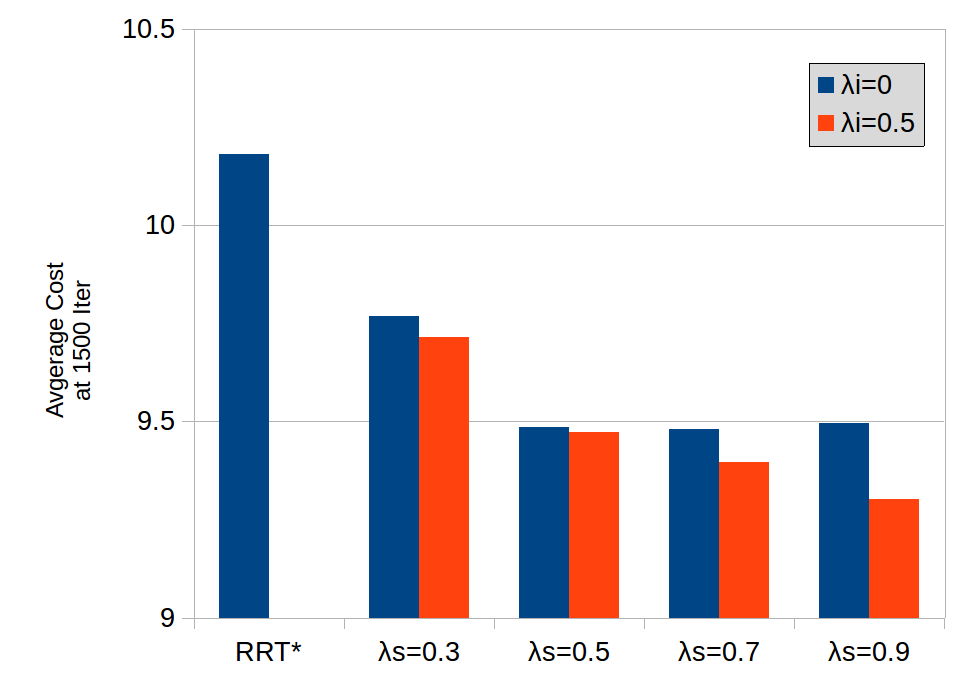}
        \caption{ }
    \end{subfigure}
        \caption{Comparison between RRT* and biased sampling w/o near-obstacle node selection bias: (a) Success Rate, (b) Cost}
    \label{fig:addimp}
\end{figure}

Next, to verify whether selecting the most important $x_{fail}$ for sampling would further improve the performance, for each sampling bias $\lambda_s$, selection bias $\lambda_i=0.5$ is added to RRT*-LDV. The rate of successfully finding the narrow passages and the cost after 1500 iterations are presented in Fig. \ref{fig:addimp} (a) and (b) respectively. The new weighted sampling strategy significantly increases the chances of detecting narrow passages and provides faster convergence. 
\begin{figure}[b]%
    \centering
    \includegraphics[width=0.48\textwidth]{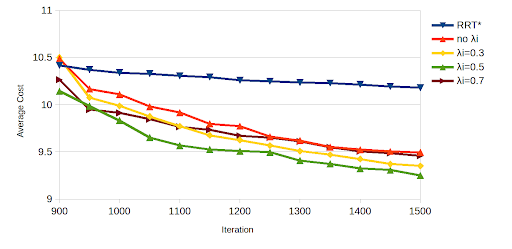}
    \caption{Sampling around the most important near-obstacle node with increasing selection bias $\lambda_i$. Sampling bias $\lambda_s=0.9$}
    \label{fig:impbias}
\end{figure}
Moreover, more bias settings are experimented by setting $\lambda_s=0.9$ and varying $\lambda_i$. The cost result shown in Fig. \ref{fig:impbias} implies that, with selection bias $\lambda_i$ added, the RRT*-LDV could find the narrow passage faster. On average before iteration 1250, added selection bias provides lower cost.

It is worth to mention, however, that $\lambda_i=0.5$ (green line in Fig. \ref{fig:impbias} provides the best performance. The reason is that, in the earlier iterations (900-1100), high $\lambda_i$ is able to identify the narrow passage faster, but it does not improve the path in later iterations (1300-1500). If biasing too much on sampling around those $x_{fail} \in X_{fail}$ that are near obstacles' vertices or in narrow passages in the later phase of searching, it is not able to improve the path in other areas. Since the sampling space is within the $d$-dimensional cube region around $X_{fail}$ with the steer size, other near-obstacle nodes would further improve the solution in later phase. On the other hand, low $\lambda_i$ is able to improve the solution cost in the later phase (yellow line in Fig. \ref{fig:impbias}). Therefore, $\lambda_i=0.5$ balances the earlier phase and the later phase and provides the best performance.

\subsection{5DOF Planning}
\label{highD}
\begin{figure}[t]%
    \centering
    \begin{subfigure}{0.24\textwidth}
        \centering
        \includegraphics[width=\textwidth]{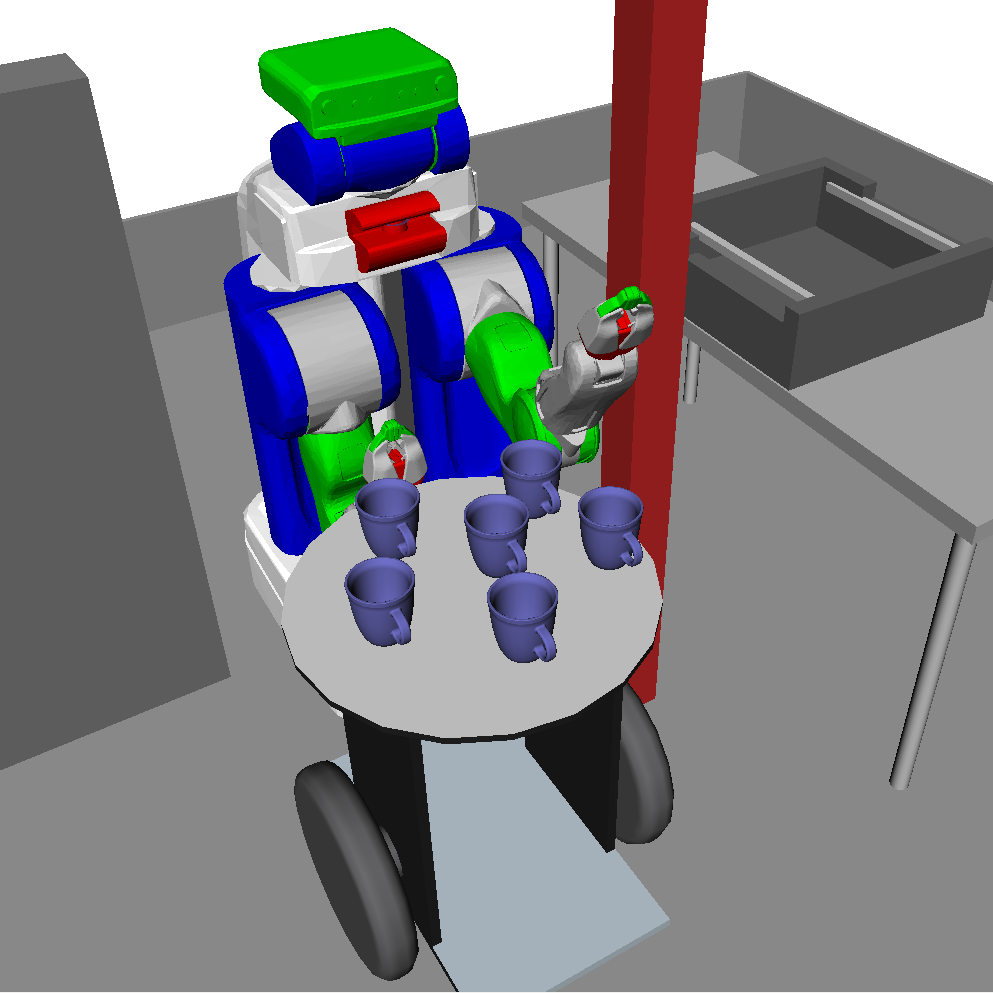}
        \caption{ }
    \end{subfigure}%
    \hfill
    \begin{subfigure}{0.24\textwidth}
        \centering
        \includegraphics[width=\textwidth]{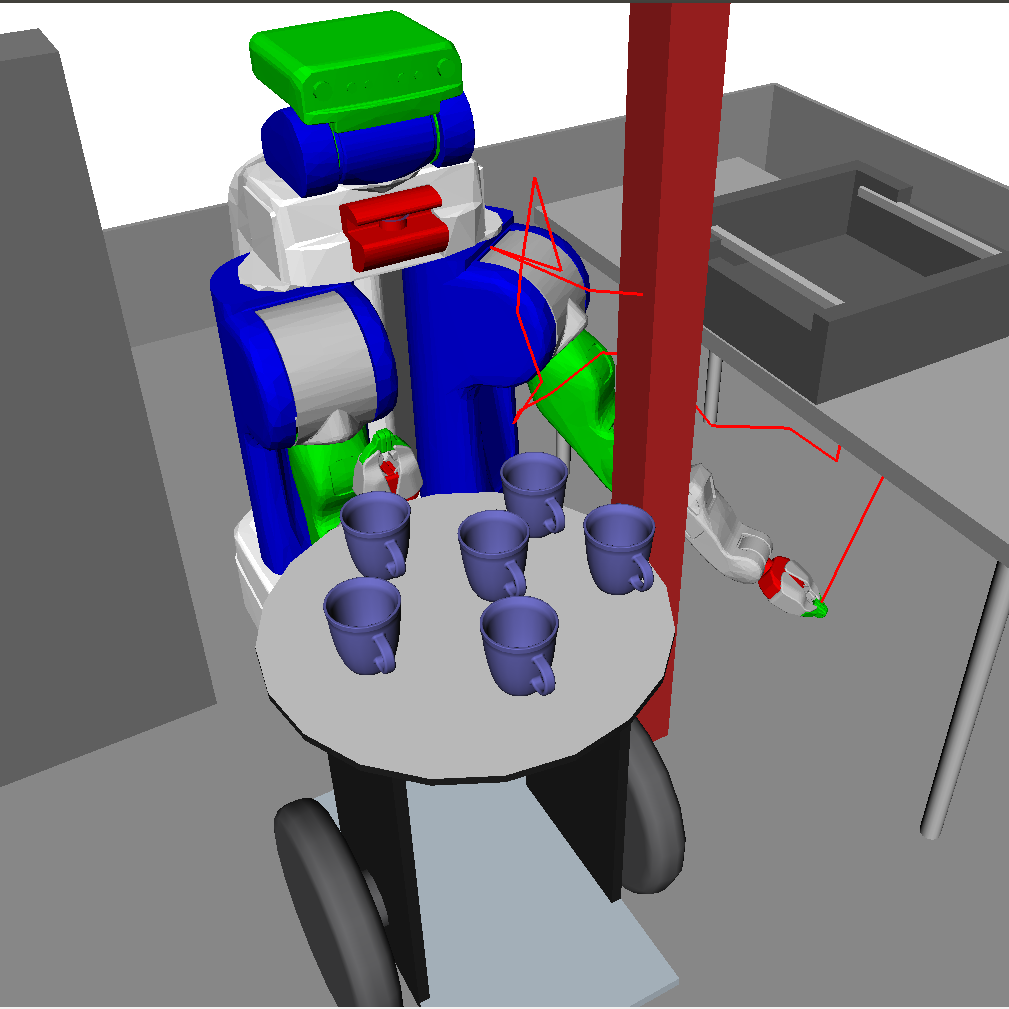}
        \caption{ }
    \end{subfigure}
        \caption{5DOF scenario. (a) the start configuration of PR2 robot. (b) Red line indicated the best path found after 15mins search.}
    \label{fig:highenv}
\end{figure}

In the high dimensional scenario, the RRT*-LDV algorithm is implemented to search for a collision-free path in the 5DOF configuration space of the robot’s arm. The algorithm starts from the current configuration of the robot’s arm shown in Fig. \ref{fig:highenv} (a) and find a path to a specified goal state as shown in Fig. \ref{fig:highenv} (b). The robot is expected to have faster convergence and have better performance under the same CPU time in high-dimensional space compared to RRT*. Cost convergence and performance under CPU time are demonstrated in the following sections.

\begin{figure}[b]%
    \centering
    \includegraphics[width=0.48\textwidth]{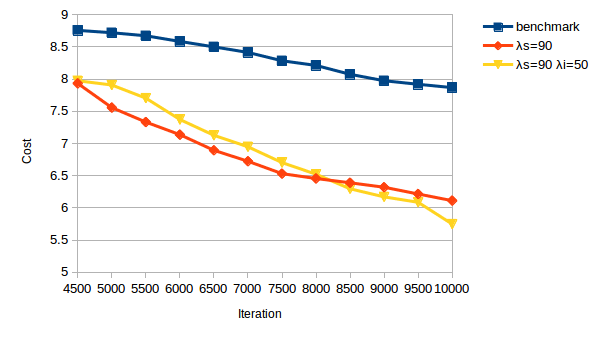}
    \caption{Comparison between RRT* and biased sampling w/o near-obstacle node selection bias in 5DOF environment}
    \label{fig:highcomp}
\end{figure}
\subsubsection{Performance Evaluation on Cost}
This experiment is run 20 times for three settings: RRT*, RRT*-LDV with sampling bias $\lambda_s =90\%$, and RRT*-LDV with sampling bias $\lambda_s=90\%$ and selection bias $\lambda_i=50\%$. The average path cost versus iteration is shown in figure \ref{fig:highcomp}. Three settings find the first solution path around 3500 iterations and end after 10000 iterations. It shows that in high-dimensional scenario, RRT*-LDV also has better performance compared with RRT* in the same iterations: RRT*-LDV already finds a shorter path ony 1000 iterations after the first solution found and has a faster convergence speed compared to RRT*. Also, without selecting based on near-obstacle node importance (red line), RRT*-LDV finds better solution in earlier phase (5000-7500), while RRT*-LDV with selection bias $\lambda_i$ provides shorter path in later phase (8500-10000). Although it is hard to explicitly see narrow passages due to DOF limits or the complexity of the environment, the cost trend implies RRT*-LDV might successfully find the narrow passage around 9500 iteration with $\lambda_i$ added.

\subsubsection{Performance Evaluation on CPU Time}
To see if RRT*-LDV has better performance under the same CPU time, another experiment is run for 15 minutes for three settings: RRT*, RRT*-LDV with sampling bias $\lambda_s =0.7$, RRT*-LDV with sampling bias $\lambda_s=90\%$ and selection bias $\lambda_i =0.3$. The result is listed in TABLE \ref{tab:maxtime}. It is shown that RRT*-LDV still outperforms the original RRT* by searching under the same time. However, due to the computational complexity in updating importance of near-obstacle nodes, selecting near-obstacle node based on importance for sampling does not provide significantly better result than randomly picking a near-obstacle node for sampling.
\begin{table}[H]
    \centering
    \begin{tabular}{|c|c|c|}
        \hline
        Algorithm Settings.     & First Path Length         &15 min Path Length   \\ \hline
        $RRT*$                  &9.67                       &7.52              \\ \hline
        $LDV,\lambda_s = 0.7$   & 9.44                      & 6.07                  \\ \hline
        $LDV,\lambda_s = 0.7,\lambda_i = 0.3$ & 9.20        & 6.06                  \\ \hline
    \end{tabular}
    \caption{Averaged path length for each case. The experiments are run on an Ubuntu 17.04}
    \label{tab:maxtime} 
\end{table}

\section{Conclusions}
\label{conclusion}
In this paper, we propose a novel algorithm accelerated RRT* by Local Directional Visibility as a variant of RRT* and it has faster convergence behavior compared to RRT*. This algorithm introduces a key property "Local Directional Visibility" of each node. During the search, the directional visibility is calculated for each node as local knowledge and near-obstacle nodes are collected as environment information in a set. After the first solution is found, RRT*-LDV performs biased sampling around these near-obstacle nodes and the probability of selecting a near-obstacle node is weighted by their importance. By running the simulation on 2DOF and 5DOF scenarios, RRT*-LDV is able to converge faster to the optimal path and finds narrow passages more easily than RRT* in complex environments.

\textit{Future works:} Since the sampling strategy biasing toward obstacles' vertices and narrow passages would help the planar to find better solution and converge to the optimal faster in the earlier searching phase, it would be desired if the sampling strategy could be biased toward the free space of the current best solution path in the later searching phase. Therefore, adaptive sampling bias and important near-obstacle node selection bias should be investigated. There are several ways of achieving so: (1) selecting a near-obstacle node for sampling by a probabilistic distribution based on importance instead of always selecting the most importance one, (2) studying a strategy of providing high selection bias at earlier phase and low selection bias at later phase under various environment, and (3) varying the weight between visibility and the number of nodes nearby when calculating importance. To make RRT*-LDV more effective, a decision on pruning certain near-obstacle nodes from the set could be made if searching around these nodes does not progress the convergence.

\section*{Acknowledgements}
We would like to thank Prof. Dmitry Berenson for providing an amazing motion planning course. We learned not only the key concepts of various historical and classical planning algorithms but also the current state-of-art algorithms related to motion planning. In addition, the opportunity of reading research paper and presenting them in conference style is so much appreciated.

\bibliographystyle{IEEEtran}
\bibliography{references}

\begin{thebibliography}{10}
\providecommand{\url}[1]{#1}
\csname url@samestyle\endcsname
\providecommand{\newblock}{\relax}
\providecommand{\bibinfo}[2]{#2}
\providecommand{\BIBentrySTDinterwordspacing}{\spaceskip=0pt\relax}
\providecommand{\BIBentryALTinterwordstretchfactor}{4}
\providecommand{\BIBentryALTinterwordspacing}{\spaceskip=\fontdimen2\font plus
\BIBentryALTinterwordstretchfactor\fontdimen3\font minus
  \fontdimen4\font\relax}
\providecommand{\BIBforeignlanguage}[2]{{%
\expandafter\ifx\csname l@#1\endcsname\relax
\typeout{** WARNING: IEEEtran.bst: No hyphenation pattern has been}%
\typeout{** loaded for the language `#1'. Using the pattern for}%
\typeout{** the default language instead.}%
\else
\language=\csname l@#1\endcsname
\fi
#2}}
\providecommand{\BIBdecl}{\relax}
\BIBdecl

\bibitem{RRT}
S.~M. Lavalle, ``Rapidly-exploring random trees: A new tool for path
  planning,'' 1998.

\bibitem{PRM}
L.~E. {Kavraki}, P.~{Svestka}, J.~{Latombe}, and M.~H. {Overmars},
  ``Probabilistic roadmaps for path planning in high-dimensional configuration
  spaces,'' \emph{IEEE Transactions on Robotics and Automation}, vol.~12,
  no.~4, pp. 566--580, 1996.

\bibitem{EST}
D.~Hsu, J.~C. Latombe, and R.~Motwani, ``Path planning in expansive
  configuration spaces.'' \emph{International Journal of Computational Geometry
  and Applications}, vol.~9, no. 4-5, p. 495–512, 1999.

\bibitem{smooth}
K.~Hauser and V.~Ng-Thow-Hing, ``Fast smoothing of manipulator trajectories
  using optimal bounded-acceleration shortcuts.'' \emph{IEEE International
  Conference on Robotics and Automation}, pp. 2493 -- 2498, 2010.

\bibitem{RRT*}
S.~Karaman and E.~Frazzoli, ``Sampling-based algorithms for optimal motion
  planning,'' \emph{International Journal of Robotics Research}, vol. 30.,
  no.~7, pp. 846--894, 2011.

\bibitem{AOP}
A.~Perez, S.~Karaman, A.~Shkolnik, E.~Frazzoli, S.~Teller, and M.~Walter,
  ``Asymptotically-optimal path planning for manipulation using incremental
  sampling-based algorithms,'' \emph{IEEE/RSJ Int’l Conf. on Intelligent
  Robots and Systems (IROS)}, vol. 302., p. 4307–4313, 2011.

\bibitem{hRRT}
C.~Urmson and R.~Simmons, ``Approaches for heuristically biasing rrt growth,''
  \emph{IROS}, vol.~2, p. 1178–1183, 2003.

\bibitem{fbias}
S.~Kiesel, E.~Burns, and W.~Ruml, ``Abstraction-guided sampling for motion
  planning,'' \emph{SoCS}, 2012.

\bibitem{RRM}
R.~Alterovitz, S.~Patil, and A.~Derbakova, ``Rapidly-exploring roadmaps:
  Weighing exploration vs. refinement in optimal motion planning,''
  \emph{ICRA}, p. 3706–3712, 2011.

\bibitem{OB}
S.~. X. T. . J.-M. L. . N.~A. Rodriguez, ``An obstacle-based rapidly-exploring
  random tree,'' \emph{IEEE International Conference on Robotics and
  Automation}, vol. 10., p. 1109, 2006.

\bibitem{EEB}
L.~Zhang and D.~Manocha, ``An efficient retraction-based rrt planner,''
  \emph{Robotics and Automation, 2008. ICRA 2008. IEEE International Conference
  on, IEEE}, vol. 302., 2008.

\bibitem{Informed}
J.~D. Gammell, S.~S. Srinivasa, and T.~D. Barfoot, ``Informed rrt*: Optimal
  sampling-based path planning focused via direct sampling of an admissible
  ellipsoidal heuristic,'' \emph{IEEE/RSJ International Conference on
  Intelligent Robots and Systems}, pp. 2997--3004, 2014.

\bibitem{ATD}
L.~Jaillet, A.~Yershova, S.~La~Valle, and T.~Siméon, ``Adaptive tuning of the
  sampling domain for dynamic-domain rrts,'' \emph{Intelligent Robots and
  Systems}, vol. 302., p. 2851–2856, 2005.

\bibitem{BIT}
J.~D. {Gammell}, S.~S. {Srinivasa}, and T.~D. {Barfoot}, ``Batch informed trees
  (bit*): Sampling-based optimal planning via the heuristically guided search
  of implicit random geometric graphs,'' \emph{IEEE International Conference on
  Robotics and Automation (ICRA)}, pp. 3067--3074, 2015.

\bibitem{RABIT}
S.~Choudhury, J.~D. Gammell, T.~D. Barfoot, S.~S. Srinivasa, and S.~Scherer,
  ``Regionally accelerated batch informed trees (rabit*): A framework to
  integrate local information into optimal path planning,'' \emph{IEEE
  International Conference on Robotics and Automation (ICRA)}, pp. 4207--4214,
  2016.

\bibitem{Smart}
F.~Islam, J.~Nasir, U.~Malik, Y.~Ayaz, and O.~Hasan, ``Rrt*-smart: Rapid
  convergence implementation of rrt* towards optimal solution,''
  \emph{Proceedings of IEEE International Conference on Mechatronics and
  Automation}, p. 1651–1656, 2012.

\bibitem{LSD}
M.~P. Brian~Ichter, James~Harrison, ``Learning sampling distributions for robot
  motion planning,'' \emph{International Conference on Robotics and
  Automation}, 2017.

\bibitem{SL}
Z.~X., L.~F., Z.~C., and F.~J., ``Self-learning rrt* algorithm for mobile robot
  motion planning in complex environments,'' \emph{Advances in Intelligent
  Systems and Computing}, vol. 302., 2016.

\end{thebibliography}

\appendices

\end{document}